\documentclass[runningheads]{llncs}

\usepackage{eccv}

\usepackage{eccvabbrv}

\usepackage{graphicx}
\usepackage{booktabs}

\usepackage[accsupp]{axessibility}  

\usepackage{hyperref}

\usepackage{orcidlink}
\usepackage{wrapfig}
\usepackage{bbm}
\usepackage{multirow}
\usepackage{upgreek}
\usepackage{xcolor}         
\usepackage{colortbl}
\definecolor{Gray}{gray}{0.95}
\newcommand{\gr}[1]{{\textcolor{gray}{#1}}}
\usepackage{arydshln}
\begin{document}

\title{WPS-SAM: Towards Weakly-Supervised Part Segmentation with Foundation Models} 

\titlerunning{Towards Weakly-Supervised Part Segmentation with Foundation Models}

\author{Xin-Jian Wu\inst{1,2},
Ruisong Zhang\inst{1,2},
Jie Qin\inst{1,2},
Shijie Ma\inst{1,2},
Cheng-Lin Liu\inst{1,2}\thanks{Corresponding author.}
}

\authorrunning{X. Wu et al.}

\institute{State Key Laboratory of Multimodal Artificial Intelligence Systems, Institute of Automation, Chinese Academy of Sciences \and
School of Artificial Intelligence, University of Chinese Academy of Sciences \\
\email{\{wuxinjian2020, zhangruisong2019, qinjie2019, mashijie2021\}@ia.ac.cn, \{liucl\}@nlpr.ia.ac.cn}
}

\maketitle

\begin{abstract}
    Segmenting and recognizing diverse object parts is crucial in computer vision and robotics. Despite significant progress in object segmentation, part-level segmentation remains underexplored due to complex boundaries and scarce annotated data.
    To address this, we propose a novel \textbf{W}eakly-supervised \textbf{P}art \textbf{S}egmentation (\textbf{WPS}) setting and an approach called \textbf{WPS-SAM}, built on the large-scale pre-trained vision foundation model, Segment Anything Model (SAM). 
    WPS-SAM is an end-to-end framework designed to extract prompt tokens directly from images and perform pixel-level segmentation of part regions. During its training phase, it only uses weakly supervised labels in the form of bounding boxes or points.
    Extensive experiments demonstrate that, through exploiting the rich knowledge embedded in pre-trained foundation models, WPS-SAM outperforms other segmentation models trained with pixel-level strong annotations. 
    Specifically, WPS-SAM achieves 68.93\% mIOU and 79.53\% mACC on the PartImageNet dataset, surpassing state-of-the-art fully supervised methods by approximately 4\% in terms of mIOU. 
\end{abstract}

\section{Introduction}
\label{sec:intro}

\begin{figure}[t]
  \centering
  \includegraphics[width=0.95\linewidth]{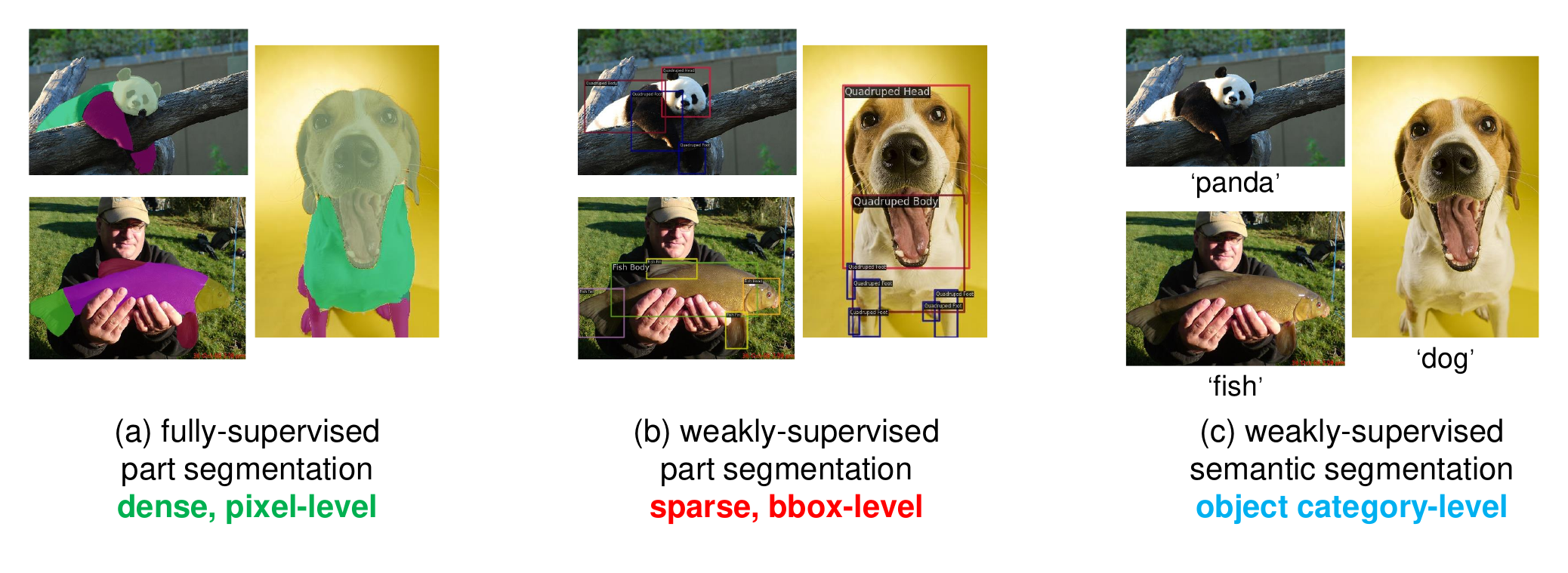}
    \caption{\textbf{Illustration of the training data comparison: (a) fully-supervised part segmentation task, (b) proposed WPS task, and (c) WSSS task.} Our approach significantly alleviates the burden of data annotation compared to fully-supervised methods, while outperforming WSSS methods in finer-grained tasks.}
  \label{fig:0}
\end{figure}

Recognizing objects and decomposing them into meaningful semantic parts is an inherent capability of human visual perception. This capability is important for human interaction with objects in the physical environment. For instance, when we type on the keyboard, it is necessary to accurately identify the boundaries of keys, as well as understand and discern the functions associated with each key. In perception and reasoning, we frequently deduce the whole from the observation of the parts of an object. This aligns with the conjecture advanced by cognitive psychologists~\cite{biederman1987recognition,lake2015human}, suggesting that hierarchical representations of objects are constructed in a bottom-up manner. Hence, developing a vision system capable of part-level segmentation is crucial and promises significant advantages across various applications in computer vision and robotics.

As one of the fundamental tasks in computer vision, object-level semantic segmentation has been extensively studied and has made significant progress~\cite{long2015fully,chen2017rethinking,xie2021segformer,cheng2021per,MIR-2022-06-191}. However, part-level segmentation presents additional challenges. Parts often possess intricate structures, characterized by complex boundaries and variations in appearance. Moreover, compared to object segmentation, there is a scarcity of pixel-level annotated data available for training part segmentation models. This scarcity has led to the emergence of various self-supervised and unsupervised part segmentation methods in the early stages~\cite{hung2019scops, liu2021unsupervised, thewlis2017unsupervised}. However, these methods often exhibit lower performance and are only applicable to a limited range of object categories.

To tackle the aforementioned challenges, we propose a novel task called Weakly-supervised Part Segmentation (WPS) in this paper, as illustrated in Figure~\ref{fig:0} (b). 
WPS enables our model to utilize cost-effective annotations, such as points or bounding boxes of parts during the training phase, resulting in pixel-level segmentation of part regions.
Compared with existing weakly-supervised semantic segmentation (WSSS) tasks that primarily focus on object-level segmentation with image-level labels~\cite{rong2023boundary,kweon2023weakly}, WPS specifically addresses the more fine-grained task of part-level semantic segmentation using sparse bounding box-level annotations. Our approach strikes a superior balance between the cost of annotation and the performance of part segmentation.

\begin{figure*}[t]
  \centering
  \includegraphics[width=\linewidth]{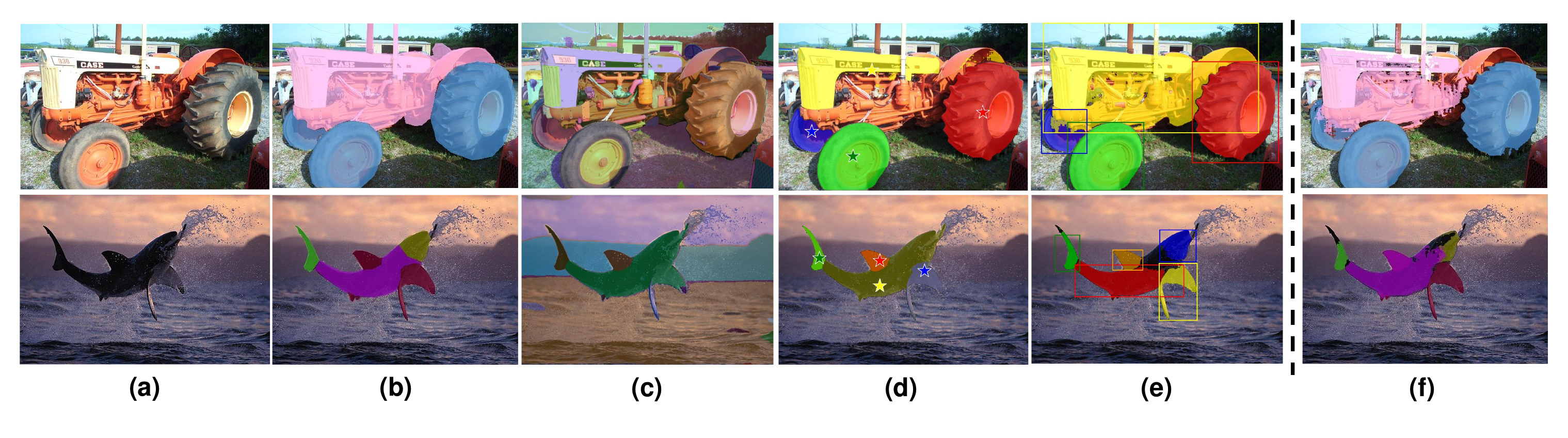}
  \caption{\small{\textbf{Visualizations of the segmentation results using pre-trained SAM directly under different modes and employing our method.} Each color represents a unique category. (a) Original images. (b) Ground truths of part segmentation. (c) The "everything" mode of SAM without prompts, segments all elements without considering the characteristics of objects and parts. (d) Segmentation results under points-form prompts, which may either miss or over-segment certain parts. (e) Segmentation results with bounding boxes prompts, achieving superior part segmentation results. (f) High-quality segmentation results of the proposed WPS-SAM method without requiring manual provision of prompts.}}
  \label{fig:1}
\end{figure*}

Empowered by extensive pre-training, vision foundation models like CLIP~\cite{radford2021learning}, ALIGN~\cite{jia2021scaling}, and DINOv2~\cite{oquab2023dinov2} have exhibited substantial potential in vision understanding, and assisting computer vision tasks, such as enabling cross-modal learning, zero-shot learning, and weakly-supervised learning. Notably, the Segment Anything Model (SAM)~\cite{kirillov2023segany} has recently emerged, showcasing remarkable segmentation capabilities on unseen objects when dealing with a wide range of images and objects. This provides an opportunity for the part segmentation field. However, SAM operates as an interactive framework, requiring a high-quality prior prompt (such as a point, box, mask, or text) alongside the input image to generate instance segmentation results. Additionally, it performs category-agnostic segmentation, as shown in Figure~\ref{fig:1}. These characteristics hinder SAM directly applied to autonomous part segmentation.

To better exploit the enriched knowledge within foundational models and explore their potential ability in the part segmentation task, we propose a novel end-to-end approach, called \textbf{WPS-SAM}, which can automate prompts generation to enhance the capabilities of SAM. 
Unlike most other segmentation methods, the remarkable feature of our approach is that it does not require pixel-level mask annotations. Instead, it solely relies on bounding box or point-level annotations, paired with their respective class labels. The former is used to guide the learning of the prompter, while the latter compensates for the insufficiency of SAM in lacking the category semantics. This helps part segmentation tasks, especially in the situation of limited annotated data. 
Specifically, we introduce a prompter based on the feature maps extracted by SAM. This prompter autonomously learns the prompt tokens corresponding to different parts in the input image, thereby replacing the previous interactive framework that relied on sequential manual guidance.

In our experiments on different part segmentation datasets, the proposed approach demonstrates superior performance over state-of-the-art methods for object and part segmentation, even though the proposed model relies on only weakly-supervised forms of annotation. For instance, WPS-SAM achieves 68.93\% mIOU and 79.53\% mACC on the PartImageNet dataset, surpassing state-of-the-art fully supervised methods by 3.69\% and 0.69\%, respectively.
In summary, we make the following contributions in this paper:
\begin{enumerate}
    \item[$\bullet$] We introduce a novel task: weakly-supervised part segmentation (WPS), aiming to find a balance between the annotation cost and the segmentation performance in the task of part semantic segmentation.
    \item[$\bullet$] By exploring the potential of the foundation model, we propose a part segmentation model (WPS-SAM) that can automatically learn part prompts, making the proposed model an end-to-end framework.
    \item[$\bullet$] Experimental results show that our WPS-SAM significantly outperforms state-of-the-art part segmentation methods which are trained with fully supervised pixel-level labels.
\end{enumerate}

\section{Related Work}
\label{sec:related work}

\subsection{Part Segmentation}

Modeling objects in terms of constituent parts has been a persistent challenge in computer vision, with a well-established and extensive research history in this domain. Beginning with the inception of Pictorial Structure~\cite{fischler1973representation} introduced in the early 1970s, numerous methods~\cite{fei2006one, girshick2011object, yuille1992feature, felzenszwalb2005pictorial, weber2000unsupervised, zhu2007stochastic, felzenszwalb2009object,chen2014detect} have been introduced to explicitly model the parts and their spatial relationships within the entire object.
The deformable part model (DPM)~\cite{felzenszwalb2009object} was once considered the most classic work in the field of object detection. 
These models collectively emphasize that object-part models provide rich representations and enhance the interpretability of prediction.

With the advancement of technology, there's a growing need for a more fine-grained understanding and segmentation of objects at the part level. In the early stages of the deep learning era, the progress of data-driven part segmentation research has been hindered by the absence of extensive datasets containing corresponding part-level mask annotations. Consequently, several self-supervised and unsupervised part segmentation methods~\cite{hung2019scops, liu2021unsupervised, thewlis2017unsupervised} emerged. However, these methods exhibit lower performance and are limited to specific classes. 
Recently, the introduction of datasets incorporating part annotations for common objects, exemplified by PartImageNet~\cite{he2022partimagenet} and PACO~\cite{ramanathan2023paco}, highlights the growing academic attention on this task, prompting further related research~\cite{he2023compositor, pan2023towards, wei2023ov}. Nonetheless, these methods still rely on pixel-level annotations, incurring high acquisition costs. In this paper, the WPS-SAM we proposed alleviates the model's reliance on strong supervision labels while achieving satisfactory performance.

\subsection{Weakly-Supervised Semantic Segmentation}      
Existing WSSS methods can be roughly grouped into single-stage and multi-stage techniques. The single-stage methods~\cite{pathak2015constrained,tang2018regularized,araslanov2020single,ru2022learning,zhang2020reliability} aim to train an end-to-end segmentation models using image-labels. The integration of classification and segmentation in single-stage approaches poses challenges for further optimization of the segmentation model, leading to suboptimal performance in current methods.
On the other hand, the common pipeline of multi-stage methods~\cite{ahn2018learning,kweon2021unlocking,rong2023boundary,kweon2023weakly} is to utilize CAMs as initial seed areas to generate pseudo-labels, and then use them to train a segmentation model, which achieves better performance. To improve the quality of pseudo labels initially generated by CAMs, various methods including adversarial erasing~\cite{kweon2021unlocking,lee2021anti,kweon2023weakly,wei2017object},  saliency guidance~\cite{lee2021railroad,yao2020saliency}, affinity learning~\cite{ahn2018learning,fan2020cian}, contrast learning~\cite{du2022weakly,zhou2022regional} and boundary-aware~\cite{li2022towards,rong2023boundary} techniques have been proposed. 

In recent developments, there a new trend~\cite{chen2023segment,lin2023clip,sun2023alternative,yang2024foundation} is to take advantage of pre-trained large foundation models~\cite{radford2021learning,kirillov2023segany} to generate high-quality pseudo labels. Although these methods have achieved state-of-the-art performance in the task, they still fall short of the performance exhibited by fully supervised methods. In addition to the conventional setting of WSSS task, several methods~\cite{dai2015boxsup,xie2021learning,khoreva2017simple,kulharia2020box2seg,song2019box,li2023sim} introduce the bounding box annotations to generate proper segmentation masks. Compared with these methods, our approach focuses on a more fine-grained task of part-level segmentation using boxes and learns prompts to activate the potential of SAM, eliminating the need for generating pseudo-labels, and achieving better performance than other fully supervised methods.

\subsection{Vision Foundation Models}

Leveraging extensive pre-training, vision foundational models have attained remarkable success in the field of computer vision. Motivated by the principles of masked language modeling~\cite{devlin2018bert, liu2019roberta} in natural language processing, MAE~\cite{he2022masked} adopts an asymmetric encoder-decoder structure and employs masked image inpainting to efficiently train scalable vision Transformer models~\cite{dosovitskiy2020image}. MAE demonstrates exceptional fine-tuning performance across a range of downstream tasks. CLIP~\cite{radford2021learning} and ALIGN~\cite{jia2021scaling} learn image representations from scratch using over a hundred million image-text pairs, demonstrating remarkable zero-shot image classification capabilities. 

While most foundation models are designed to extract accessible knowledge from freely available data, the recent SAM method~\cite{kirillov2023segany} establishes a data engine involving collaborative model development alongside dataset annotation through model-in-the-loop processes. Thanks to pre-training on 1 billion masks and 11 million images, SAM showcases impressive zero-shot, task-agnostic segmentation performance. This has spurred a series of studies~\cite{zou2023segment,chen2023semantic,wang2023seggpt,liu2023matcher,chen2023rsprompter,liu2023samm, li2023semantic, MIR-2023-11-249} applying SAM to specific downstream tasks. Motivated by these advancements, our study explores SAM's robust general segmentation capabilities in weakly-supervised part segmentation, aiming to introduce a fresh perspective to readers.

\section{Methodology}
\label{sec:methodology}
In this section, we analyze the characteristics of SAM on the part segmentation task. Then we introduce the proposed WPS-SAM for achieving end-to-end part segmentation while relying on weakly-supervised labels only during training. The training and inference techniques will be presented in sequence.

\subsection{A Closer Look at SAM}
\label{subsec: a closer look at SAM}

\begin{figure}[t]
  \centering
  \resizebox{0.48\textwidth}{!}{
    \begin{minipage}[b]{0.48\textwidth}
            \captionof{table}{\small{
                The preliminary experimental results of SAM on the PartImageNet \textit{val} set with various types of prompts and backbones, which reflect the performance \textbf{upper bound} of our method.}}
            \begin{tabular}{ccccc}
                \toprule
                Backbone & Label & mIoU(\%) & mACC(\%)\\
                \midrule
                \multirow{2}{*}{ViT-B~\cite{dosovitskiy2021an}} & point & 72.64 & 90.63 \\ 
                                                                & bbox  & 91.14 & 98.19 \\ \hline
                \multirow{2}{*}{ViT-L~\cite{dosovitskiy2021an}} & point & 71.14 & 89.67 \\ 
                                                                & bbox  & 91.62 & 98.34 \\ \hline                          
                \multirow{2}{*}{ViT-H~\cite{dosovitskiy2021an}} & point & 71.45 & 89.95 \\ 
                                                                & bbox  & 91.80 & 98.43 \\           
            \bottomrule
            \end{tabular}
            \label{tab:0}
    \end{minipage}
    }
\hfill    
    \begin{minipage}{0.48\linewidth}
        \vskip -0.7in
          \centering
          \includegraphics[width=1\linewidth]{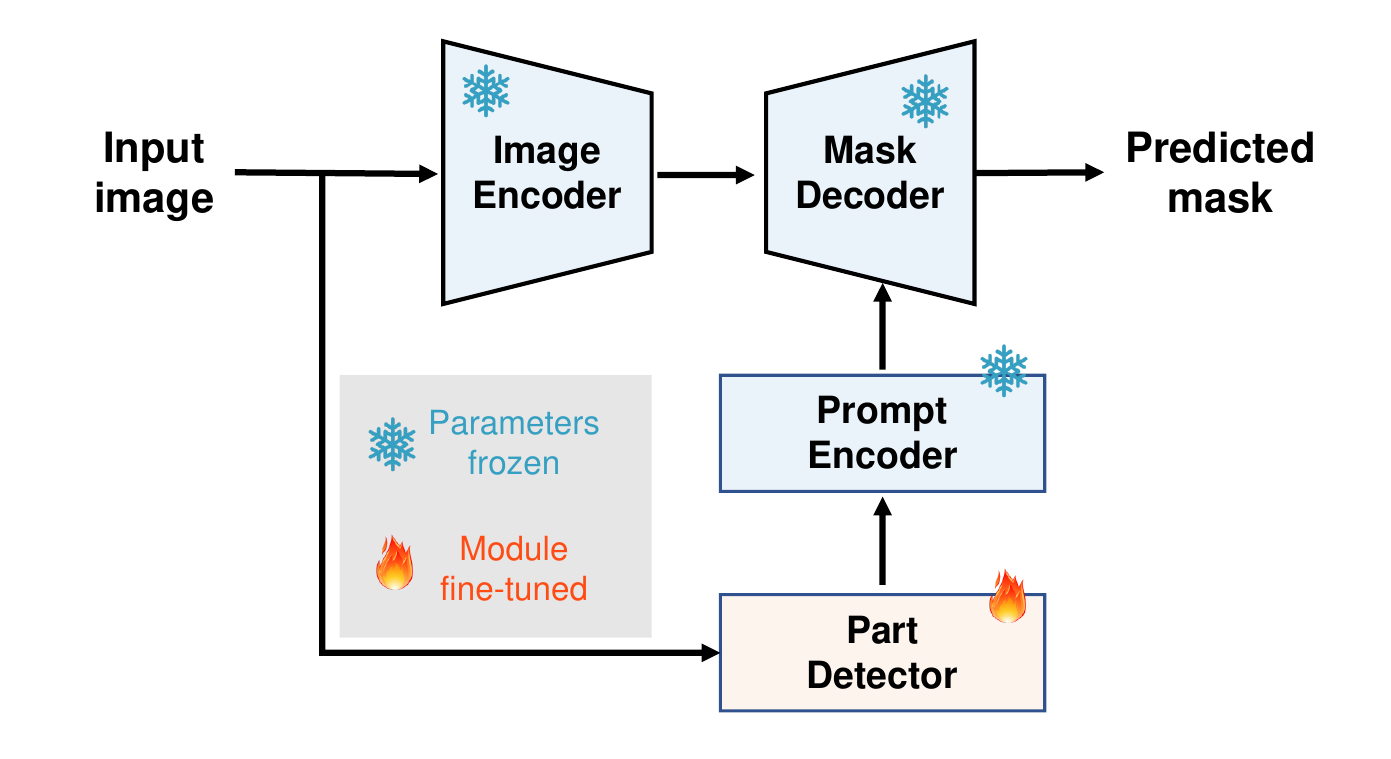}
          \caption{\small{\textbf{Schematic diagram of the trivial Det-SAM.} We argue that a simple combination of a detector and SAM is not the most optimal solution. }}
          \label{fig:5}
    \end{minipage}
\end{figure}

As shown in Figure~\ref{fig:1}, with high-quality prompts, SAM can yield satisfactory results in part segmentation even without any fine-tuning.
To validate this, we conducted preliminary experiments on the PartImageNet dataset with the pre-trained SAM. Specifically, we use annotated bounding boxes and the center points as prompts for SAM input, and the resulting part segmentation outcomes are presented in Table~\ref{tab:0}. This impressive performance strongly reflects the capabilities of the foundation model in this task. However, manually providing prior prompts is cumbersome in practical applications. 

To explore the ability of SAM more efficiently, it is desired to learn or generate prompts automatically. A trivial approach, named Det-SAM in this paper, involves training a dedicated detector initially to detect the bounding boxes of parts. Subsequently, these detected bounding boxes are used as prompts for SAM input, as illustrated in Figure~\ref{fig:5}. However, we find that this simple strategy does not obtain satisfactory performance because the part detector is imperfect and does not cooperate with the segmentation module very well, more detailed analysis is shown in Section~\ref{subsec:ablation study}. To address this issue and further exploit the rich visual information within the foundation model, we propose an end-to-end strategy that directly generates prompts by utilizing the feature maps extracted by the SAM image encoder, which is more efficient and achieves better performance than the above Det-SAM.

\subsection{End-to-end Part Segmentation Framework}

\begin{figure*}[!t]
  \centering
  \includegraphics[width=1.0\linewidth]{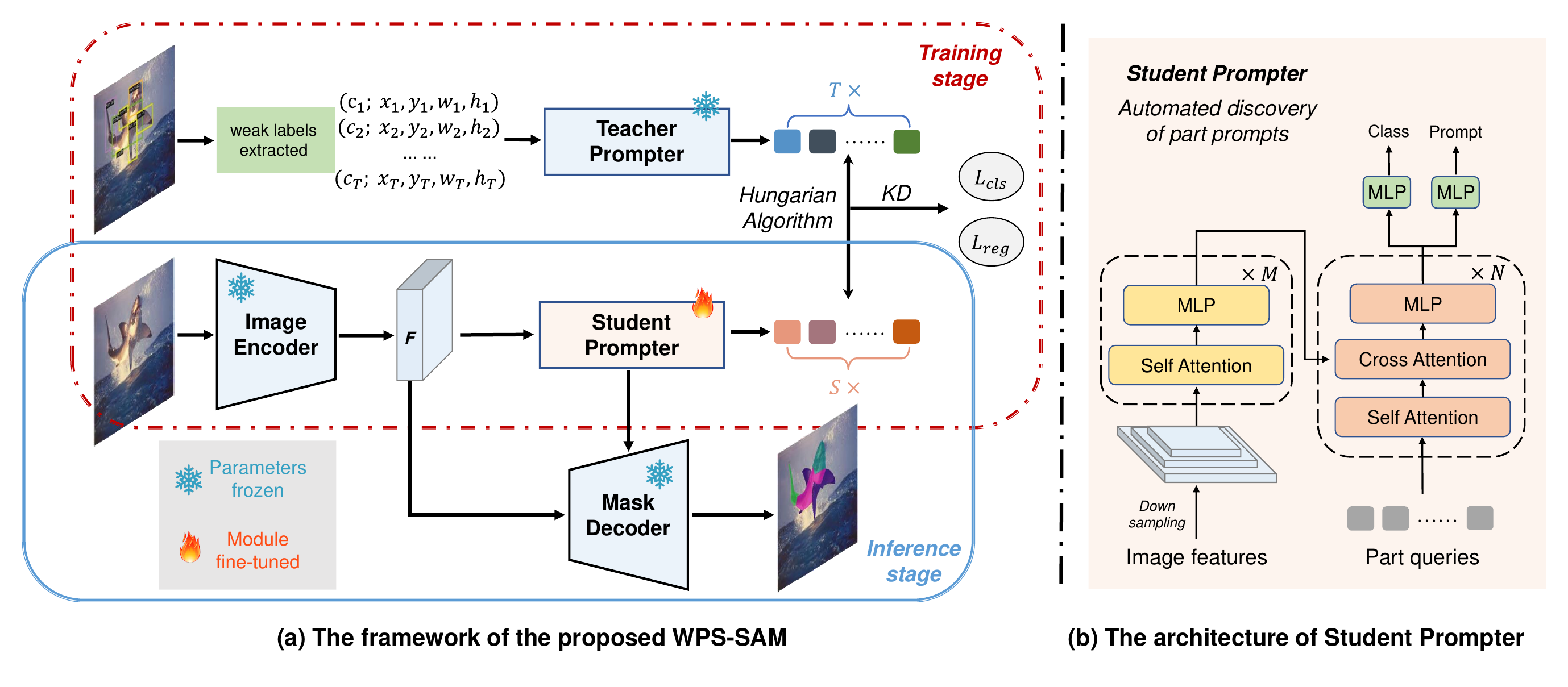}
  \caption{\small{\textbf{An overview of the proposed framework WPS-SAM}, accomplishing part segmentation in an end-to-end manner while relying solely on cost-effective weak labels during training. The modules with frozen parameters in the figure come from the pre-trained SAM~\cite{kirillov2023segany}. Additionally, the utilized student prompts are derived from a lightweight query-based Transformer architecture.}}
  \label{fig:2}
\end{figure*}

\noindent\textbf{Overview.} As depicted in Figure~\ref{fig:2} (a), the overall WPS-SAM architecture is concise and intuitive, consisting of three main modules: the image encoder for feature extraction, the student prompter for prompts generation, and the mask decoder for mask prediction. The entire pipeline is expressed as follows:
\begin{align}
\begin{split}
    F_{\text{img}} &= \Phi_{\text{i-enc}}(\mathcal{I}) \\
    T_{\text{prompt}} &= \Phi_{\text{prompter}}(F_{\text{img}}) \\
    \mathcal{M}_{\text{out}} &= \Phi_{\text{m-dec}}(F_{\text{img}}, T_{\text{prompt}})\\ \label{eq:overall}
\end{split} 
\end{align}
where $\mathcal{I} \in \mathbb{R}^{1024 \times 1024 \times 3}$ represents the resized original image, $F_{\text{img}} \in \mathbb{R}^{64 \times 64 \times 256}$ denotes the embedded image features, and $T_{\text{prompt}} \in \mathbb{R}^{S \times 256}$ signifies the prompt tokens encoded by our proposed prompter $\Phi_{\text{prompter}}$, $\mathcal{M}_{out} \in \mathbb{R}^{S \times 1024 \times 1024}$ corresponds to the predicted part masks, where $S$ is the number of part queries.

\noindent\textbf{Query-based prompter.} As illustrated in Figure~\ref{fig:2} (b), the proposed prompter is primarily based on a transformer encoder-decoder structure, incorporating several CNN layers as well as two feed-forward networks.

Considering the large size of the feature map extracted by the image encoder, which imposes a significant computational burden on subsequent modules, we initially designed two layers of $3\times3$ convolutional layers to implement downsampling and further information fusion.

The encoder plays a crucial role in information fusion and extracting higher-level semantic features from the image features. Subsequently, the decoder is responsible for converting a set of learnable queries into the output embedding by interacting via cross-attention with the semantic features. These output embeddings are then independently predicted into prompt tokens with corresponding categories by the MLPs. These prompt tokens corresponding to the parts in the image are precisely what is needed to accomplish the part segmentation task.

We aim to augment the capabilities of the SAM prompter, enabling it to autonomously derive semantic prompts based on the current input. To preserve the knowledge already acquired by SAM and minimize computational costs, We freeze the parameters in image encoder $\Phi_{\text{i-enc}}$ and mask decoder $\Phi_{\text{m-dec}}$ of SAM. Then we learn the proposed student prompter based on the knowledge distillation technology~\cite{hinton2015distilling}. Specifically, we employ the original SAM's prompt encoder as the teacher network, utilizing it to receive weakly supervised data annotations, \textit{e.g.}, points or bounding boxes. Subsequently, we use its outputs to supervise the training of the student prompter $\Phi_{\text{prompter}}$, which takes image features embedded by $\Phi_{\text{i-enc}}$ as input. 

Finally, the trained $\Phi_{\text{prompter}}$ can replace the original prompter, enabling the end-to-end part segmentation task.

\subsection{Training and Inference} 

\noindent\textbf{Training.} The query-based prompter generates a fixed-size set of $S$ predictions, whereas the actual number of parts in the current image $T$ is variable. One of the primary challenges in training is to evaluate the predicted parts (category, prompt embedding) of the pseudo-labels generated by the teacher prompter. Our loss function establishes an optimal bipartite matching between predictions and pseudo-labels and subsequently optimizes parts-specific losses, which simultaneously considers category prediction and the regression task for teacher prompt embeddings.

Let \(y\) represent the supervised label set of parts, and \(\hat{y} = \{\hat{y}_i\}_{i=1}^{S}\) denote the set of \(S\) predictions. Assuming that \(S\) is greater than the number of parts in the image, we treat \(y\) as a set of size \(S\) padded with \(\varnothing\) (no parts).

To establish a bipartite matching between these two sets, we seek a permutation of \(S\) elements \(\sigma \in \mathcal{\sigma}_{S}\) with the minimum cost:
\begin{equation}
\label{eq:matching}
    \hat{\sigma} = \arg\min_{\sigma\in\mathcal{\sigma}_{S}} \sum_{i}^{S} C_{m}(y_i, \hat{y}_{\sigma(i)}),
\end{equation}
where \(C_{m}({y_i, \hat{y}_{\sigma(i)}})\) represents the pairwise matching cost between the ground truth \(y_i\) and a prediction with index \(\sigma(i)\). The optimal assignment is efficiently computed using the Hungarian algorithm, as outlined in prior works~\cite{carion2020end,stewart2016end}.

The matching cost considers both the class prediction and the similarity between predicted prompts and pseudo-labels generated by the teacher prompter. Each element \(i\) in the supervised label set can be represented as \(y_i = (c_i, p_i)\), where \(c_i\) is the target class label (which may be \(\varnothing\)), and \(p_i\) is the prompt embedding generated by the teacher prompter based on weak labels.
For the prediction with index $\sigma(i)$, we define the probability of class $c_i$ as $\hat{p}_{\sigma(i)}(c_i)$ and the predicted prompts as $\hat{p}_{\sigma(i)}$. With these notations, we define the matching cost as follows:
\begin{equation}
    C_{m}(y_i, \hat{y}_{\sigma(i)}) = \mathbbm{1}_{\{c_i\neq\varnothing\}}(-\alpha \hat{p}_{\sigma(i)}(c_i) + \beta L_2({p_{i}, \hat{p}_{\sigma(i)}})),
\end{equation}
where $\alpha$ and $\beta$ are weight coefficients used to balance the two types of costs.

Once each predicted instance is paired with its corresponding ground truth under the optimal assignment $\hat{\sigma}$ computed in step~\ref{eq:matching}, we can compute the loss function to optimize the parameters as described below:
\begin{align}
\begin{split}
    \mathcal{L}(y, \hat{y}) = \frac{1}{S} \sum_{i}^{S}(\lambda_{cls} \mathcal{L}_{\text{cls}}^i + \mathbbm{1}_{\{c_i\neq\varnothing\}} \lambda_{reg}\mathcal{L}_{\text{reg}}^i),
    \label{eq:loss}
\end{split} 
\end{align}
where $\mathcal{L}_{\text{cls}}$ denotes the cross-entropy loss computed between the predicted category and the target (contain the category of $\varnothing$), while $\mathcal{L}_{reg}$ signifies the smooth $L_1$ loss between the predicted prompt embeddings and the matched teacher prompt embeddings.

\noindent\textbf{Inference.} In the inference phase, we no longer rely on the Hungarian matching process, because of the lack of supervised teacher labels. Instead, we directly retain the prompt tokens corresponding to the foreground (specific part categories) and discard those related to the background (empty category $\varnothing$) based on the predictions from the classification head in the student prompter. 

\section{Experiments}
\label{sec:experments}

\subsection{Setup}
This paper aims to achieve weakly-supervised part semantic segmentation. To validate the concept of our problem and the effectiveness of our approaches, we use the following datasets. During the model training, we exclusively utilize part annotations in the form of points or bounding boxes, reserving pixel-level masks for performance evaluation during inference. We adopt standard evaluation metrics for semantic segmentation, \textit{i.e.}, mean Intersection over Union (mIoU), and mean Pixel Accuracy (mACC).

\noindent\textbf{PartImageNet}~\cite{he2022partimagenet}. This dataset is a large, high-quality dataset with part segmentation annotations following the COCO style. It comprises 158 classes from ImageNet~\cite{deng2009imagenet}, totaling 24,080 images. The classes are organized into 11 super-categories, and the part splits are designed based on these super-categories, totaling 40 part categories.

\noindent\textbf{PASCAL-Part}~\cite{chen2014detect}. The original Pascal Part dataset offers part annotations for 20 classes from Pascal VOC~\cite{everingham2010pascal}, encompassing a total of 193 part categories. The training and validation sets comprise 10,103 images, while the testing set contains 9,637 images. Following the setting of ~\cite{he2023compositor}, we only consider the 16 classes that have part-level annotations and ignore the rest. We manually merge the provided labels
to a higher-level definition of parts (e.g. "eyes", "ears", "nose", etc. can be merged into a single "head" part) since the original parts are too fine-grained.

\noindent\textbf{Implementation details.} During training, we maintain the image size at 1024 $\times$ 1024, consistent with SAM, and refrain from applying additional augmentations. We only train the parameters of the introduced prompter while freezing the parameters of other parts of the network. The student prompter includes a downsampling layer with two layers of $3\times3$ convolutional kernels with a stride of 2. Both the encoder and decoder layers in the student prompter are set to 6 layers. And we set $\alpha=10.0,\ \beta=1.0,\ \lambda_{\text{cls}}=5.0,\ \lambda_{\text{reg}}=20.0$ respectively. We utilize the Adam optimizer with a learning rate of $1e-4$ for prompter training and set the batch size to 8 on each GPU. The total training epochs amount to 150. Our source code is publicly available at \url{https://github.com/xjwu1024/WPS-SAM}.

\subsection{Main Results}
\label{subsec:main results}

\begin{table*}[t]
    \small
    \centering
    \caption{Comparisons between WPS-SAM and classical fully-supervised segmentation models, as well as state-of-the-art weakly supervised semantic segmentation (WSSS) methods on the \textbf{PartImageNet} \textit{val} set.}
    \vskip -0.15in
    \resizebox{0.95\textwidth}{!}{
    \begin{tabular}{c l l| l| c| c| c} 
    \toprule
    & Method & Venue & Backbone & Annotation & mIoU(\%) & mACC(\%) \\ \hline
    \multirow{9}{*}{\rotatebox{90}{\textit{Fully-Supervised}}}
    &  $\text{Deeplab v3+}$~\cite{chen2018encoder} & \gr{ECCV'18} & & & 60.57 & 71.07 \\
    &  $\text{MaskFormer}$~\cite{cheng2021per} &  & & & 60.34 & 72.75 \\
    &  $\text{MaskFormer-Dual}$ & \multirow{-2}{*}{\gr{NeurIPS'21}} & &  & 58.02 & 70.42 \\
    &  $\text{Compositor}$~\cite{he2023compositor} & \gr{CVPR'23} & \multirow{-4}{*}{ResNet-50~\cite{he2016deep}} & \multirow{-4}{*}{mask} & \text{61.44} & \text{73.41} \\
    \cline{2-7}
    &  $\text{SegFormer}$~\cite{xie2021segformer} & \gr{NeurIPS'21} & MiT-B2~\cite{xie2021segformer} & & 61.97 & 73.77 \\
    &  $\text{MaskFormer}$~\cite{cheng2021per} & & & & 63.96 & 77.37 \\
    &  $\text{MaskFormer-Dual}$ & \multirow{-2}{*}{\gr{NeurIPS'21}} & &  & 61.69 & 75.64 \\
    &  $\text{Compositor}$~\cite{he2023compositor} & \gr{CVPR'23} & \multirow{-3}{*}{Swin-T~\cite{liu2021swin}} & & \text{64.64} & \text{78.31} \\
    &  $\text{MaskFormer}$~\cite{cheng2021per} &{\gr{NeurIPS'21}}& Swin-B~\cite{liu2021swin}  & \multirow{-5}{*}{mask}  & 65.24 & 78.84 \\
    \hline
    \hline
    &  \textbf{WPS-SAM} & This Work & ViT-B~\cite{dosovitskiy2021an} & & \textbf{68.93} & \textbf{79.53} \\
    \multirow{3}{*}{\rotatebox{90}{\textit{WSSS}}}
    &  $\text{SIM}$~\cite{li2023sim} & \gr{CVPR'23} & ResNet-101~\cite{he2016deep} & \multirow{-2}{*}{bbox} & \text{49.51} & \text{63.27} \\    
    \cline{2-7}
    &  $\text{BECO}$~\cite{rong2023boundary} & \gr{CVPR'23} & &  & \text{42.37} & \text{53.07} \\
    &  $\text{FMA-WSSS}$~\cite{yang2024foundation} & \gr{WACV'24} & \multirow{-2}{*}{ResNet-101~\cite{he2016deep}} & \multirow{-2}{*}{category} & \text{56.74} & \text{68.07} \\    \hline   
    \bottomrule
    \end{tabular}}
    \label{tab:1}
    \vskip -0.2in
\end{table*}

\noindent\textbf{Comparisons with other methods.} 
We conducted comparisons between our proposed WPS-SAM and classical fully-supervised segmentation models, as well as state-of-the-art weakly supervised semantic segmentation (WSSS) methods. The fully-supervised segmentation models contain CNN-based semantic segmentation models~\cite{chen2018encoder,cheng2021per}, as well as more advanced Transformer-based architectures~\cite{cheng2021per, xie2021segformer}. Additionally, we compared our approach with the latest state-of-the-art research~\cite{he2023compositor} about part segmentation. 
Even though only sparse annotations at the bounding box level are introduced in our method, the experimental results in Table~\ref{tab:1} and Table~\ref{tab:2} demonstrate that our method outperforms other methods that use stronger pixel-level dense annotations. On another hand, our approach achieves significantly better performance than existing WSSS methods without the need for cumbersome steps such as generating pseudo-labels. Moreover, leveraging a pre-trained visual foundation model in our approach results in a significant portion of parameters being frozen. This significantly reduces the number of trainable parameters in comparison to alternative methods which leads to a decrease in computational costs during the training process.

Table~\ref{tab:1} presents a summary of our experimental results on PartImageNet. Utilizing ViT-B as the backbone, our proposed method, WPS-SAM, achieves 68.93\% mIoU and 79.53\% mACC. These results surpass those of fully-supervised methods of comparable scale by approximately 4\% and state-of-the-art WSSS method by around 12\% in terms of mIOU.

\begin{table*}[t]
    \small
    \centering
    \caption{Comparisons between WPS-SAM and classical fully-supervised segmentation models, as well as state-of-the-art weakly supervised semantic segmentation (WSSS) methods on the \textbf{PASCAL-Part} \textit{val} set.}
    \vskip -0.1in
    \resizebox{0.95\textwidth}{!}{
    \begin{tabular}{c l l| l| c| c| c} 
    \toprule
    & Method & Venue & Backbone & Annotation & mIoU(\%) & mACC(\%) \\ \hline
    \multirow{7}{*}{\rotatebox{90}{\textit{Fully-Supervised}}}
    &  $\text{MaskFormer}$~\cite{cheng2021per} &  & & & 47.61 & 58.59 \\
    &  $\text{MaskFormer-Dual}$ & \multirow{-2}{*}{\gr{NeurIPS'21}} & & & 46.60 & 57.96 \\
    &  $\text{Compositor}$~\cite{he2023compositor} & \gr{CVPR'23} & \multirow{-3}{*}{ResNet-50~\cite{he2016deep}} & \multirow{-3}{*}{mask} & \text{48.01} & \text{58.83} \\ 
    \cline{2-7}
    &  $\text{MaskFormer}$~\cite{cheng2021per} & & & & 55.42 & 67.21 \\
    &  $\text{MaskFormer-Dual}$ &\multirow{-2}{*}{\gr{NeurIPS'21}} & & & 54.21 & 66.42 \\
    &  $\text{Compositor}$~\cite{he2023compositor} & \gr{CVPR'23} & \multirow{-3}{*}{Swin-T~\cite{liu2021swin}} & & \text{55.92} & \text{67.63} \\
    &  $\text{MaskFormer}$~\cite{cheng2021per} &{\gr{NeurIPS'21}}& Swin-B~\cite{liu2021swin} & \multirow{-4}{*}{mask} & 56.83 & 68.46 \\
    \hline \hline
    &  $\textbf{WPS-SAM}$ & This Work &ViT-B~\cite{dosovitskiy2021an} & & \textbf{60.49} & \textbf{71.25} \\
    \multirow{3}{*}{\rotatebox{90}{\textit{WSSS}}}
    &  $\text{SIM}$~\cite{li2023sim} & \gr{CVPR'23} & ResNet-101~\cite{he2016deep} & \multirow{-2}{*}{bbox} & \text{37.82} & \text{49.56} \\    
    \cline{2-7}
    &  $\text{BECO}$~\cite{rong2023boundary} & \gr{CVPR'23} & &  & \text{34.53} & \text{46.66} \\
    &  $\text{FMA-WSSS}$~\cite{yang2024foundation} & \gr{WACV'24} & \multirow{-2}{*}{ResNet-101~\cite{he2016deep}} & \multirow{-2}{*}{category} & \text{42.21} & \text{54.13} \\    \hline
    \bottomrule
    \end{tabular}}
    \label{tab:2}
    \vskip -0.15in
\end{table*}

\begin{figure*}[!t]
  \centering
  \includegraphics[width=1.0\linewidth]{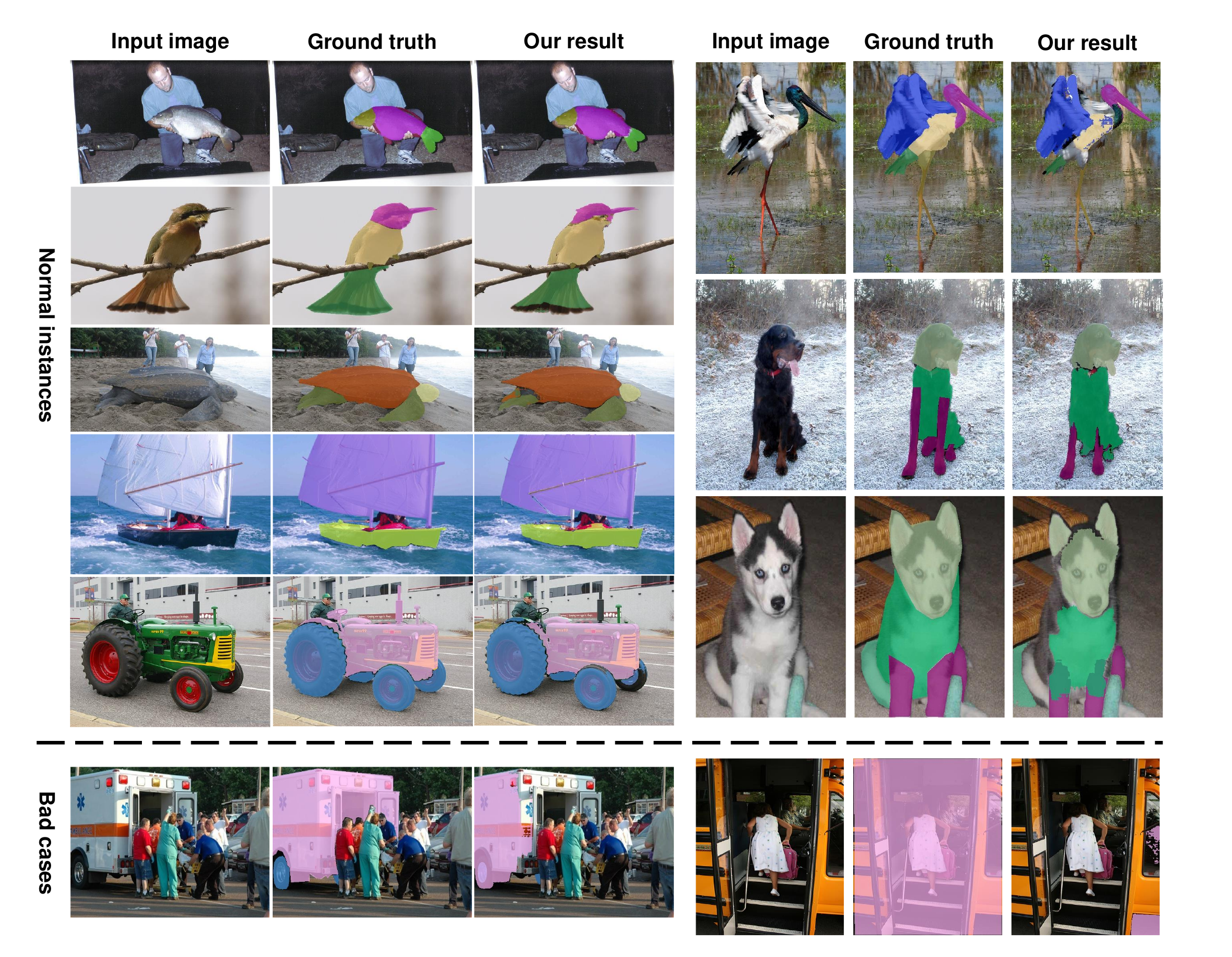}
  \vskip -0.15in
  \caption{\small{\textbf{Qualitative results for part segmentation on PartImageNet generated by WPS-SAM with ViT-B backbone}.} The masks with different colors correspond to different part categories. These high-quality part predictions demonstrate the feasibility of part segmentation in diverse real-world scenarios and underscore the effectiveness of our proposed approach. We present segmentation results for the majority of typical examples (above the dashed line) and showcase a few bad cases (below the dashed line) where parts are occluded or lack clear semantic information.}
  \label{fig:3}
\end{figure*}

Table~\ref{tab:2} presents our experimental results on Pascal-Part. Notably, in contrast to the images in PartImageNet, which typically contain a single object, Pascal-Part scenes are more complex, involving multiple objects. This complexity results in a performance decline for various methods, however, our approach maintains a significant advantage. With ViT-B as the backbone, WPS-SAM achieves 60.49\% mIoU and 71.25\% mACC. In short, we demonstrate that our approach attains stronger performance by leveraging the rich knowledge within SAM, even with weaker annotations.

\noindent\textbf{Visualizations of part segmentation results.} 
In order to offer a more intuitive understanding of the proposed WPS-SAM, we conducted the following visualization analysis, as illustrated in Figure~\ref{fig:3}. Despite being trained exclusively on sparse, box-level part annotations, our model demonstrates a remarkable capability to generate high-quality part segmentation results that exhibit close alignment with the ground truths. This noteworthy accomplishment underscores the robustness and efficacy of WPS-SAM in capturing intricate part details based on minimal annotation information.
Moreover, it is crucial to acknowledge the presence of challenging cases in specific scenarios, such as situations where the target is obscured or semantic information is unclear. This also provides a direction for us to further refine our approach.

\subsection{Ablation Study}
\label{subsec:ablation study}
In this section, we conduct a series of experiments on PartImageNet to investigate the importance of each
component and parameter setting in our proposed method. Unless otherwise specified, the ViT-B is used as the default backbone of the image encoder.

\begin{table*}[!t]
\small
  \centering
  \caption{Performance comparison of different frameworks, where \textbf{\textit{Backbone}} refers to the part detector's underlying architecture.}
  \begin{tabular}{cccccc}
    \toprule
    Framework & Detector & Backbone &  mAP(\%) & mIoU(\%) & mACC(\%)\\
    \midrule
    \multirow{4}{*}{Det-SAM} & Faster-RCNN~\cite{NIPS2015_14bfa6bb} & \multirow{2}{*}{ResNet-50} & 35.0 & 63.7 & 74.8 \\
     & DETR~\cite{carion2020end} & & 34.2 & 61.4 & 70.2\\
     \cline{2-6} 
     & Faster-RCNN~\cite{NIPS2015_14bfa6bb} & \multirow{2}{*}{SAM-Encoder} & 24.2 & 54.6 & 65.2 \\
     & DETR~\cite{carion2020end} & & 26.3 & 56.3 & 68.6\\ \hline 
      \textbf{WPS-SAM} & - & - & - & \textbf{68.9} & \textbf{79.5} \\
    \bottomrule
  \end{tabular}
  \label{tab:3}
\end{table*}

\begin{figure*}[!t]
  \small
  \centering
  \includegraphics[width=1.0\linewidth]{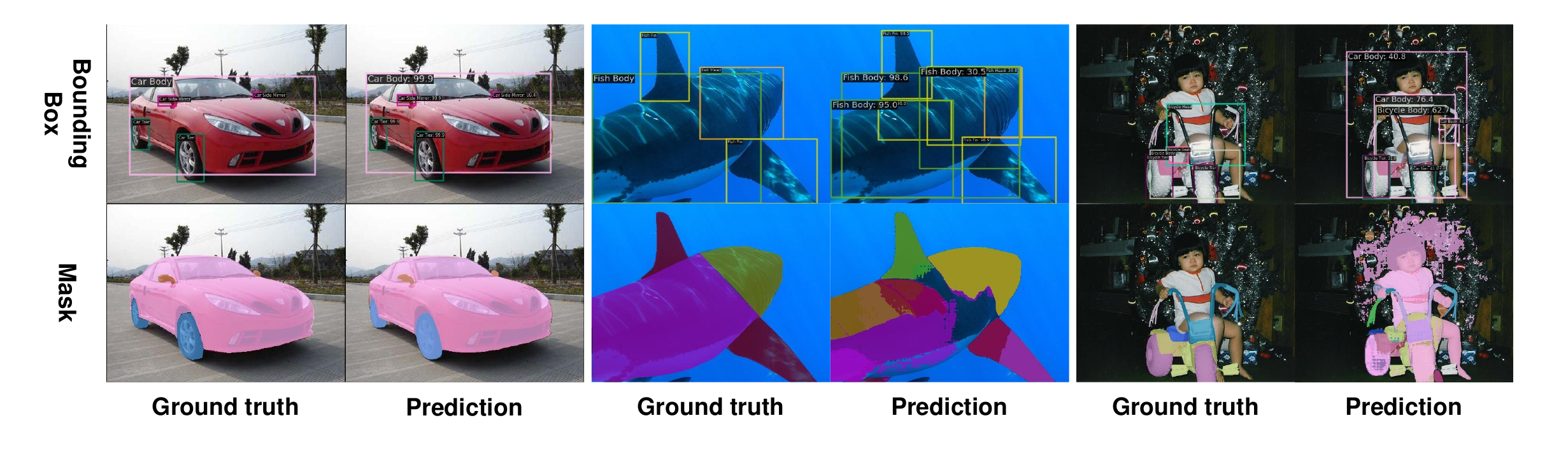}
  \caption{\textbf{The visualization results of Det-SAM}, with Faster-RCNN as a detector, which includes both box-level and mask-level information, clearly demonstrate the significant impact of part detection on the performance of part segmentation.}
  \label{fig:4}
\end{figure*}

\noindent\textbf{Det-SAM \textit{V.S.} WPS-SAM.} We evaluate the effectiveness of our framework by considering Det-SAM, which involves training a dedicated part detector independently and connecting it to the prompter of SAM, as discussed in Section~\ref{subsec: a closer look at SAM}. We utilize two classic detectors, Faster R-CNN~\cite{NIPS2015_14bfa6bb} and DETR~\cite{carion2020end}, as part detectors and compare their performance with our approach. The results are presented in Table~\ref{tab:3}. Det-SAM achieves performance comparable to existing methods but significantly lower than our proposed framework, further emphasizing the effectiveness of our approach. 
We attribute the superior performance of our method to two main factors. Firstly, using a pre-trained image feature extractor, which has been trained on large-scale datasets, enables our framework to leverage richer visual information. Secondly, we employ high-dimensional prompt embeddings as teacher labels for training the student prompter, which encapsulates more informative cues than the more traditional bounding box representations. 
Furthermore, we conducted experiments to evaluate the performance of Det-SAM by replacing the backbone of the part detectors with the pre-trained SAM-encoder. The results showed a decrease in performance, which could be attributed to compatibility issues between the frameworks and suboptimal adjustments of training details.
Figure~\ref{fig:4} illustrates the qualitative segmentation results obtained using Det-SAM. It is evident that the accuracy of the part detector heavily influences the segmentation performance.

\noindent\textbf{Compared with the pseudo-label paradigm.} Based on the preliminary results presented in Table~\ref{tab:0}, we observed that SAM demonstrates the capability to generate high-quality segmentation masks. These masks can be used as pseudo-labels to train existing segmentation models, following a two-phase approach commonly employed in WSSS methods. Furthermore, we report the performance of MaskFormer trained using these pseudo-labels in Table~\ref{tab:6}. It is evident that the performance of MaskFormer trained with SAM-generated pseudo-labels is competitive with ground truth masks supervised. However, it still falls slightly behind our WPS-SAM, highlighting the advantages of SAM in the segmentation task compared to other models.

\begin{table}[!t]
\small
  \centering
  \caption{Performance comparison of MaskFormer with ground truth mask, pseudo-labels, and WPS-SAM on \textbf{PartImageNet} \textit{val} set.}
  \resizebox{0.8\textwidth}{!}{%
  \begin{tabular}{ccccc}
    \toprule
    Method & Backbone & Supervision & mIoU(\%) & mACC(\%)\\
    \midrule
    \multirow{2}{*}{MaskFormer~\cite{cheng2021per}} & \multirow{2}{*}{Swin-B} & gt-mask & 65.24 & 78.84\\
    & & bbox $\rightarrow$ pseudo-mask & 56.83 & 69.57\\
    \textbf{WPS-SAM} & ViT-B & bbox & \textbf{68.93} & \textbf{79.53}\\
    \bottomrule
    \label{tab:6}
  \end{tabular}}
\end{table}

\begin{wrapfigure}[7]{r}{0.45\textwidth} 
    \centering
    \begin{minipage}[t]{\linewidth} 
        \captionof{table}{\small{Performance comparison on points and boxes supervised.}}
        \begin{tabular}{ccc}
        \toprule
        supervision & mIoU(\%) & mACC(\%)\\
        \midrule
        point & 52.28 & 66.85  \\
        bbox & \textbf{68.93} & \textbf{79.53} \\
        \bottomrule
        \end{tabular}
        \label{tab:4}
    \end{minipage}
\end{wrapfigure}
\noindent\textbf{Boxes or Centers?} We also explore another widely used form of weak labeling, specifically \textit{center point} annotation, to assess the performance of WPS-SAM. Table~\ref{tab:4} demonstrates that under the point annotation form, the performance of WPS-SAM is significantly inferior to that of the bounding box annotation. We attribute this disparity to the fact that while point annotation can locate parts, it fails to capture the shape of the parts. In contrast, bounding boxes can to some extent reflect this information, resulting in stronger performance, as illustrated in Figure~\ref{fig:1}.

\begin{wrapfigure}[9]{r}{0.45\textwidth} 
    \centering
    \begin{minipage}[t]{\linewidth} 
        \captionof{table}{\small{Ablation study on the number of queries.}}
        \begin{tabular}{ccc}
        \toprule
        \#queries & mIoU(\%) & mACC(\%)\\
        \midrule
        \textbf{25} & \textbf{68.93} & \textbf{79.53}\\
        50 & 64.74 &  75.07\\
        75 & 62.93 &  72.40\\
        100 & 62.71 & 71.80\\
        \bottomrule
        \end{tabular}
        \label{tab:5}
    \end{minipage}
\end{wrapfigure}
\noindent\textbf{Number of queries.} Considering that the number of parts in a single image is usually not too large (generally not exceeding 20), so we set the number of queries in our proposed prompter to 25. Table~\ref{tab:5} presents the performance of the model under different query quantities. When increasing the parameter count to 50/75/100, we noticed a slight decrease in performance. This decrease may be attributed to the additional convergence burden placed on the model due to the larger parameter count.

Due to space limitations, more discussions can be seen in the appendix.

\section{Conclusion}
\label{sec:conclusion}

In this paper, we introduce WPS-SAM, a novel weakly-supervised prompt learning method for part segmentation. WPS-SAM strikes a balance between annotation cost and segmentation performance by automatically learning part prompts, without manual guidance. Leveraging pre-trained foundation models, WPS-SAM outperforms other segmentation methods relying on strong-supervised annotations, achieving remarkable results with a 68.93\% mIOU and 79.53\% mACC on the PartImageNet dataset. We conduct a comprehensive analysis to highlight the advantages of our approach over Det-SAM and the pseudo-label paradigm. Additionally, we delve into an exploration of the performance of our method under different weak supervision (box and point), along with an investigation into the impact of various hyperparameters on our model. Despite the promising results, a limitation of our proposed method is its reliance on SAM, which is computationally heavy and hinders its applicability in resource-constrained environments. We suggest incorporating lightweight variants of SAM. such as EfficientSAM~\cite{xiong2023efficientsam} and FastSAM~\cite{zhao2023fast}, to mitigate this limitation.

Overall, our work offers a valuable contribution to the development of part segmentation, showcasing the effectiveness of incorporating foundation models. We hope that our method inspires new research and leads to further improvements in weakly-supervised part segmentation.

\section*{Acknowledgements}
This work has been supported by the National Natural Science Foundation of China (NSFC) grant U20A20223.

\bibliographystyle{splncs04}
\bibliography{main}

\newpage
\renewcommand{\thetable}{\Alph{table}}
\setcounter{table}{0}  
\setlength{\intextsep}{5pt} 
\appendix

\section{Ablation study on SAM fine-tuning}

To evaluate the impact of fine-tuning the SAM encoder in our framework using box-level supervision, we conducted an ablation study. The results, presented in Table~\ref{tab:A}, indicate a reduction in segmentation performance when the SAM encoder is fine-tuned. This reduction can be attributed to the coarse-grained nature of the box-level annotations, which contrasts with the typical fine-tuning practice where annotations are more precise than those used in pre-training. Additionally, fine-tuning the SAM encoder required significantly more GPU memory and increased training time.

\begin{table*}[h]
\small
  \centering
  \caption{Performance comparison of SAM Encoder: \textbf{\textit{Fine-tuning vs. Frozen}}.}
  \vspace{-0.1in}
  \begin{tabular}{ccccc}
    \toprule
         & SAM Encoder & mIoU (\%) & mACC (\%) & training time\\
        \midrule
        \multirow{2}{*}{WPS-SAM} & fine-tuning & 65.16 & 75.79 & 127h \\
            & \textbf{frozen} & \textbf{68.93} & \textbf{79.53} & 59h \\
    \bottomrule
  \end{tabular}
  \label{tab:A}
\end{table*}

\vspace{-0.2in}
\section{Ablation study on network architectures}

To investigate the impact of different network architectures, we conducted an ablation study comparing classical CNN architectures within an anchor-based framework to a query-based transformer architecture. The results, presented in Table~\ref{tab:B}, indicate a performance decline when using the anchor-based CNN approach. This decline may be attributed to the incompatibility between the CNNs and the transformer architectures when combined within the model.

\begin{table*}[!h]
\small
  \centering
  \caption{Performance comparison of different network architectures.}
  \vspace{-0.1in}
  \begin{tabular}{cccc}
    \toprule
     & architecture & mIoU (\%) & mACC (\%)\\
    \midrule
    \multirow{2}{*}{Student prompter} & anchor-based (CNN) & 57.23 & 68.11 \\
        & \textbf{query-based (Transformer)} & \textbf{68.93} & \textbf{79.53} \\
    \bottomrule
  \end{tabular}
  \label{tab:B}
\end{table*}

\vspace{-0.2in}
\section{Ablation study on hyper-parameters}

We conducted detailed ablation studies on critical hyper-parameters to understand their impact on performance. The results, presented in Table~\ref{tab:C}, provide insights into how different hyper-parameter settings affect our framework.

\begin{table*}[!h]
\small
  \centering
  \caption{Performance comparison of different hyper-parameters.}
  \begin{tabular}{ccccccc}
    \toprule
     & $\alpha$ & $\beta$ & $\lambda_{cls}$ & $\lambda_{reg}$ & mIoU (\%) & mACC (\%)\\
    \midrule
    \multirow{6}{*}{WPS-SAM} & 1 & 1 & 1 & 1 & 39.35 & 47.40\\
        & 5 & 1 & 5 & 1 & 56.94 & 69.59\\
        & 5 & 1 & 10 & 1 & 61.93 & 77.07\\
        & 5 & 10 & 10 & 1 & 67.24 & 78.49\\
        & \textbf{5} & \textbf{20} & \textbf{10} & \textbf{1} & \textbf{68.93} & \textbf{79.53}\\
        & 5 & 50 & 10 & 1 & 67.82 & 77.54\\
    \bottomrule
  \end{tabular}
  \label{tab:C}
\end{table*}

\end{document}